%% file: main.tex
\documentclass[
]{ceurart}

\usepackage{amssymb}
\usepackage{graphicx}
\usepackage{float}
\usepackage[capitalise,nameinlink]{cleveref}

\usepackage{listings}
\begin{document}

\copyrightyear{2023}
\copyrightclause{Copyright for this paper by its authors.
  Use permitted under Creative Commons License Attribution 4.0
  International (CC BY 4.0).}

\conference{CLEF 2023: Conference and Labs of the Evaluation Forum, September 18--21, 2023, Thessaloniki, Greece}

\title{Enhancing Biomedical Text Summarization and Question-Answering: On the Utility of Domain-Specific Pre-Training}

\title[mode=sub]{University of Technology Sydney participation in BioASQ Task 11b Phase B}


\author[]{Dima Galat}[%
orcid=0000-0003-3825-2142,
email=dima.galat [@] student.uts.edu.au,
url=https://github.com/dimagalat/
]
\author[]{Marian-Andrei Rizoiu}[%
orcid=0000-0003-0381-669X,
email=Marian-Andrei.Rizoiu [@] uts.edu.au,
url=https://www.rizoiu.eu/,
]
\address[]{University of Technology Sydney (UTS), Australia}


\begin{abstract}
  Biomedical summarization requires large datasets to train for text generation. We show that while transfer learning offers a viable option for addressing this challenge, an in-domain pre-training does not always offer advantages in a BioASQ summarization task. We identify a suitable model architecture and use it to show a benefit of a general-domain pre-training followed by a task-specific fine-tuning in the context of a BioASQ summarization task, leading to a novel three-step fine-tuning approach that works with only a thousand in-domain examples. Our results indicate that a Large Language Model without domain-specific pre-training can have a significant edge in some domain-specific biomedical text generation tasks. 
  \end{abstract}

\begin{keywords}
  natural language processing \sep
  biomedical summarization \sep
  biomedical question answering \sep
  transfer learning \sep
  language modeling \sep
  domain-specific pre-training \sep
  BioASQ \sep
  CEUR-WS
\end{keywords}

\maketitle

\input{sections/1-introduction-v1.tex}

\section{Related work}
We are looking for a LLM which has an architecture suitable for long-form question answering and has been trained on relevant in-domain data. There are several important model architectures and pre-training objectives used to optimize the models worth considering \cite{Radford2018, Devlin2018, Lewis2019}. 

First, lets briefly mention BERT \cite{Devlin2018} in the context of text generation, since most biomedical Transformer-based \cite{Vaswani2017} models still rely on this architecture. BERT does not have an autoregressive decoder, preventing it from generating text. Despite this fact, a well-known summarisation approach called PreSumm \cite{Liu2019b} uses this architecture by inserting additional tokens for teaching models which sentences should be included in the summary. We followed the process proposed by the authors while using a BioBERT \cite{Lee2019a} model; we first trained an extractive summariser, which did perform a little better on BioASQ data than a regular BERT trained the same way. Unfortunately, when training an abstractive summarization architecture, PreSumm \cite{Liu2019b} process uses a randomly initialised Transformer \cite{Vaswani2017} for a decoder. It appears that there is a significant mismatch between this decoder and a BioBERT \cite{Lee2019a} encoder leading to unstable abstractive fine-tuning process and poor generation outputs in our experiments. Based on these findings, we have concluded that BERT is a wrong architecture to be using for text generation tasks.


BART \cite{Lewis2019} is an architecture that uses an encoder with an auto-regressive decoder, similarly to the original Transformer \cite{Vaswani2017}. BART relies on an architecture which can be seen as generalising BERT (because it also uses a bi-directional encoder) and GPT \cite{Radford2019} (because it also uses the left-to-right decoder). This model is using a masked language modeling objective (also known as denoising) introduced by BERT \cite{Devlin2018} and adds two additional denoising objectives (token deletion and sentence permutation). Authors conduct experiments that are focused on text generation, and show that denoising objectives are particularly well-suited for summarization tasks. Because it can be easily fine-tuned directly for generation tasks, authors achieved a remarkable success on a wide range of abstractive summarization and long-form question answering problems \cite{Lewis2019}. 

BioBART \cite{Yuan2022} is a BART model pre-trained on PubMed \cite{PubMed} abstracts. Authors have reported that they have trained without one of the objectives proposed by BART, namely the sentence permutation, showing that models trained without this objective have a better performance. Overall, this is the only study that we are aware of that applies a LLM to a range of generation tasks and reports the results (another BioGPT \cite{Luo2022} study we found has not reported any numeric results on text generation problems). We are also not completely convinced that some of the results, like those reported for a BioASQ task could not be a result of a random chance, since the differences in the scores are very small and there are a few possible sources of non-determinism in training and generation procedures we discuss later in this paper. 


\section{Our Contribution}

In the biomedical domain, the majority of models we have reviewed are focused on the pre-training process, perhaps because pre-training data is readily available \cite{Alsentzer2019,Lee2019a,Gu2020,Alrowili2021,Yuan2022,Luo2022,Raza2022}. However, question answering and summarization are plagued by a lack of a large domain specific dataset for fine-tuning LLMs directly for text generation problems. More specifically, when we are looking at the biomedical text generation tasks, it's hard to find a large (and clean) sequence-to-sequence dataset for fine-tuning for a long-form question answering and summarization. BioASQ is the closest dataset currently available, however it is still a few orders of magnitude away from what we would require to fine-tune a LLM for a previously unseen generation task. Therefore, we conclude that this two-step fine-tuning process offers a limited utility for this problem.

%

Following a conventional transfer learning definition we use a task to refer to training on labeled data, seeking to transfer the knowledge from a source task and a source domain ($\mathcal{T}_S$  and $\mathcal{D}_S$) to a target task and a target domain ($\mathcal{T}_T$ and $\mathcal{D}_T$) \cite{Pan2010,Ruder2019Neural}. One of the common transfer learning scenarios involves learning the tasks sequentially, one after another; and we could also have an intermediate fine-tuning task making it a three-step fine tuning process, where a second step is only used to get a representation that is more suitable for the task of summarization in a biomedical domain. This means that an intermediate $\mathcal{T}_{inter}$ (which could be both in/out domain) should lead to a performance improvement in $\mathcal{T}_T$. This could be potentially useful, since task-domain specific data is hard to come by.

Since we need to perform text generation, a reasonable option is to train for an $\mathcal{T}_{inter}$ which teaches the model to perform this task. Unfortunately, large question answering and summarization datasets like SQUAD \cite{Rajpurkar2016} and CNN/DM \cite{Kocisky2015} have nothing to do with biomedical domain, but because we need 10-100 times more biomedical summarization data than what we have available, we believe that task-specific datasets could offer just as much value as a domain-specific pre-training. We believe that CNN/DM is the most suitable (clean, large, easily available) task-specific dataset; especially because summaries there are typically closely related to source sentences, which is also the case with the BioASQ data. Moreover, lengths of summaries are similar to those in BioASQ. Therefore, we are interested in this task, even though a newsmedia domain would likely have completely different marginal probability distributions of generated text. This approach means that in addition to sequential transfer learning (two and three step fine-tuning processes described above), models competing with a two-step fine-tuning strategy would have to also adapt for the domain difference (i.e. differences in prior and conditional distributions). Second $\mathcal{T}_{inter}$ we have considered for training is Pubmed \cite{PubMed} article-abstracts combinations. While these are not summaries in the stricter sense of the word, this is the closest domain-specific dataset that we could find, and we would like to understand if it adds useful information to a LLM.

%
%
%
%
%


\section{Models compared}

We select LLMs that reduce the amount of overall training required. 
We select a mix of domain-specific pre-training and general pre-training datasets, and we attempt different $\mathcal{T}_{inter}$s to see how well the resulting models generalize to $\mathcal{T}_T$, namely BioASQ Task 11b Phase B. Hence, the final list of LLMs we are considering are:
\begin{itemize}
	\item BART - a baseline two-step LLM (without additional fine-tuning) used to establish a baseline for a general domain model without specialized domain knowledge or $\mathcal{T}_{inter}$ fine-tuning
	\item BioBART - a biomedical two-step LLM (without fine-tuning $\mathcal{T}_{inter}$), used to establish a baseline for an in-domain model
	\item BART CNN - a baseline LLM three-step LLM with task-specific fine-tuning $\mathcal{T}_{inter}$ but without any deep domain knowledge
	\item BioBART CNN - a biomedical three-step LLM with task-specific fine-tuning $\mathcal{T}_{inter}$	
	\item BART CNN Pubmed - a general domain  three-step LLM fine-tuned for $\mathcal{T}_{inter}$ summarisation task, and then further fine-tuned on a domain-specific $\mathcal{T}_{inter}$ dataset containing Pubmed articles 
\end{itemize}

Based on the data available, we believe that these tasks and LLMs offer the greatest benefit for biomedical summarization, and we limit our selection to 5 models that will participate in the BioASQ competition. 
We are only considering large models because we want the model to analyze as much context as possible, and therefore having a large model helps to double the context length (1024 tokens vs. 512 tokens). 
We are using pre-trained BART, BioBART, and BART CNN models available via Huggingface\footnote{\url{https://huggingface.co/models}}; and we are fine-tuning BART CNN on Pubmed data and BioBART on CNN data for one epoch each (our $\mathcal{T}_{inter}$). Subsequently, all models are fine-tuned on the latest complete BioASQ 11 dataset ($\mathcal{T}_T$) for five epochs using a 10-fold cross-validation process. We empirically chose the number of training epochs to maximize the final model scores. 
We've tried training on Pubmed (with and without training on CNN), and found it beneficial when using a general-domain model. Despite this, CNN dataset is a much better $\mathcal{T}_T$ for BioASQ. Using Pubmed (summarisation) data for fine-tuning BioBART before or after CNN training didn't offer advantages (\cref{figure:pubmedf}, \cref{figure:pubmedp}) and was excluded from the top five models under consideration.

\section{Results}

\begin{figure*}[t]
  \centering
  \label{figure:median-f1}
  \includegraphics[width=0.85\textwidth,height=\textheight,keepaspectratio]{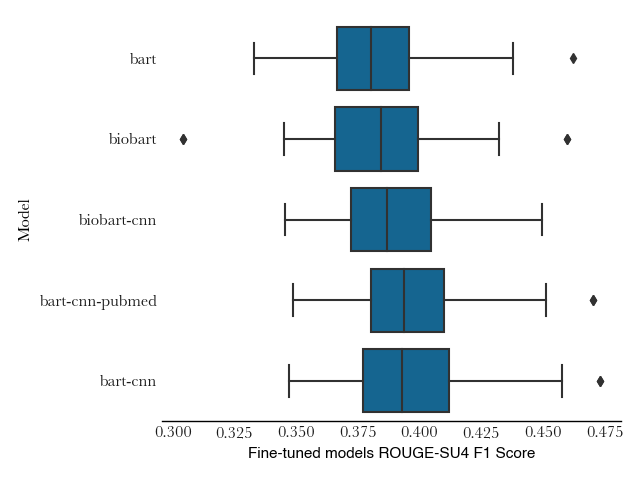}
  \caption{
    We can see that BioBART outperforms regular BART, which can be attributed to the domain-specific vocabulary; however, the same vocabulary hurts its ability to do task-specific training on the out-of-domain CNN dataset. BART CNN shows a solid performance for a model that was only fine-tuned on a small BioASQ in-domain dataset, and adding an in-domain Pubmed dataset reduces the range of scores a little, making it the best overall model.
  }
\end{figure*}

Our experiments have revealed a substantial (over 10\%) variation in ROUGE \cite{Lin2004} score results based on a simple choice of a seed parameter for cross-validation. 
This indicates that the fine-tuning process is susceptible to changes in data. 
Future studies should consider which BioASQ information snippets are passed to the model as the input for summarization training. Working with small healthcare question-answering datasets can require a more careful knowledge extraction process \cite{Liu2020}. 

We have experimented with fine-tuning for up to 10 epochs on the $\mathcal{T}_T$, and found that this problem consistently persists across a range of training scenarios. In-domain studies we have reviewed show that the generation results can often differ by a minimal margin, significantly lower than the variation in scores we have observed in cross-validation.

To our knowledge, this research is the first to draw attention to this specific problem, and we decided to overcome this by repeating the 10-fold cross-validation training process $\mathcal{T}_T$ four times using a different seed value. 
Therefore, we effectively report the average of 400 runs for each model (95\% t-test confidence interval is given in parentheses), with 100 runs for each seed choice (ten for each fold). 
We are primarily focused on SU4-F1 scores (\cref{table:meanconf-f1}) since they have been shown to correlate with human scores the best \cite{Molla2019}.
However, ROUGE is recall-oriented; therefore, we also look at Recall results separately (\cref{table:meanconf-recall}). 

\begin{table*}[]
\caption{ROUGE-SU4 F1 - Mean confidence intervals}
\label{table:meanconf-f1}
\begin{tabular}{lllll}
    \toprule
model name                  & mean  & CI 95\%        &  &  \\
    \midrule
BioBART       & 0.383 & (0.373, 0.394) &  &  \\
BART           & 0.384 & (0.376, 0.392) &  &  \\
BioBART CNN      & 0.39  & (0.382, 0.398) &  &  \\
BART CNN        & \textbf{0.396} & (0.387, 0.406) &  &  \\
BART CNN Pubmed & \textbf{0.396} & (0.386, 0.405) &  &  \\
  \bottomrule
\end{tabular}
\end{table*}

\begin{table*}[]
\caption{ROUGE-SU4 Recall - Mean confidence intervals}
\label{table:meanconf-recall}
\begin{tabular}{lllll}
    \toprule
model name      & mean  & CI 95\%        &  &  \\
    \midrule
BioBART         & 0.398 & (0.386, 0.409) &  &  \\
BART            & 0.4   & (0.392, 0.409) &  &  \\
BioBART CNN     & 0.415 & (0.407, 0.424) &  &  \\
BART CNN        & 0.42  & (0.411, 0.429) &  &  \\
BART CNN Pubmed & \textbf{0.422} & (0.41, 0.434)  &  &  \\
  \bottomrule
\end{tabular}
\end{table*}

Our experiments (\cref{figure:median-f1}) suggest that LLMs without domain-specific pre-training show a better capacity for domain-specific text generation. This becomes particularly clear when comparing BART and BioBART results before any additional task-specific fine-tuning, suggesting that BioASQ data is not as similar to Pubmed pre-training data as we would expect based on other results reported on discriminatory tasks. 
Moreover, we believe that currently a non-domain specific CNN summarization task $\mathcal{T}_{inter}$ is required to accomplish the best results on a BioASQ task. 
Adding in-domain Pubmed data improves Recall; however, Pubmed data is unsuitable for training for a summarization task from scratch. ROUGE Recall scores (\cref{figure:median-f1}) show one notable difference, BART CNN has a higher recall, whereas BART CNN Pubmed has a higher precision, likely because the Pubmed training after the task-specific training introduces a task-specific vocabulary to the model.

Overall, LLMs have established some remarkable results in various practical applications. 
However, since LLMs require task-specific datasets to train to generate text, and such domain-specific datasets are scarce, we need to find ways to overcome these challenges. 
We have presented an approach that focuses on applications of transfer learning to a domain with limited task-specific training data.

\section{Conclusion and Future Work}

In this work, we have observed that task-specific data is critical for generating text in a biomedical domain. 
Based on our experiments, models without in-domain pre-training are better at summarizing BioASQ data. 
Unfortunately, our models have achieved fairly modest automated ROUGE scores during BioASQ 11 runs, and we are waiting for the final results to determine how the models have performed overall. The generation process is non-deterministic, and while the answers generated by the models appear sensible, we need better ways to evaluate the candidates.

We have discussed how transfer learning can overcome challenges with data availability.
We see a lot of exciting possibilities for using generator models (more specifically paraphrasing, simplification, and rewriting models \cite{Xu2019}) for creating synthetic training data, as well as for providing a differentiable loss function which allows sampling a wider space of possible answers without over-penalizing exploration. 
Abstractive summarization models are trained to generate specific gold sequences, even when they start making errors in the first steps (a problem known as exposure bias \cite{Bengio2015}).
One recent improvement over BART proposes generating multiple candidates and comparing them, showing a new SOTA on several popular summarization datasets \cite{Liu2022}. 
This could address a common shortcoming of autoregressive models, leading to further performance improvements. 
Another possibility that shows a significant promise would be generating synthetic data to augment BioASQ. 
This approach has recently shown good results in machine translation \cite{Briakou2022}, and we believe it can be used for other text-generation problems.

\bibliography{main}

\appendix

\section{Additional results}
\begin{figure*}[b]
\centering
\caption{ROUGE SU4-Recall}
\label{figure:median-recall}
\includegraphics[width=0.85\textwidth,height=\textheight,keepaspectratio]{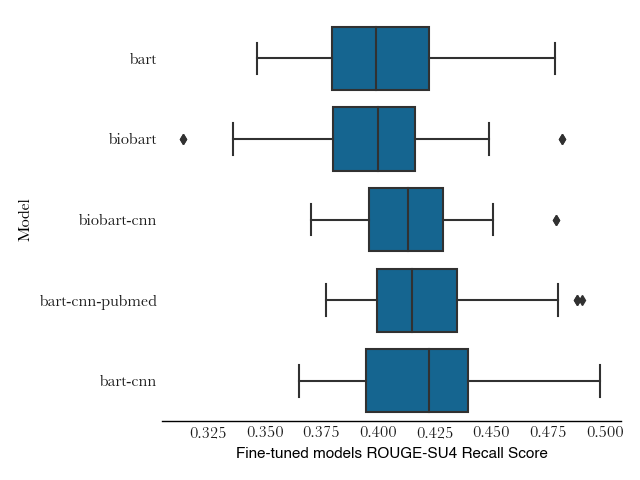}
\end{figure*}

\begin{figure*}[b]
\centering
\label{figure:pubmedf}
\caption{Biobart CNN + Pubmed: ROUGE SU4-F1}
\includegraphics[width=0.85\textwidth,height=\textheight,keepaspectratio]{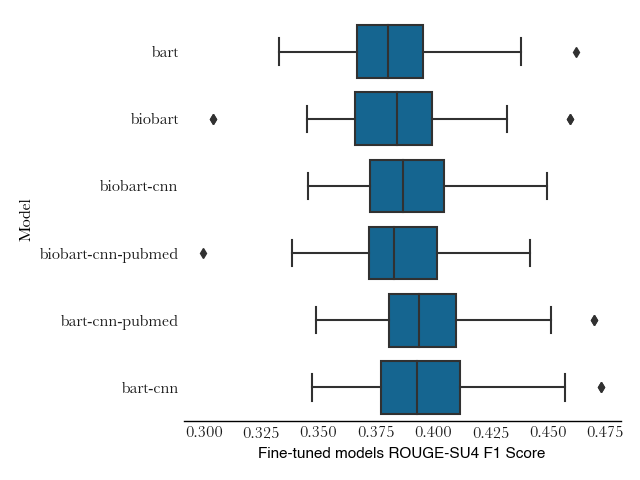}
\end{figure*}

\begin{figure*}[b]
\centering
\label{figure:pubmedp}
\caption{Biobart CNN + Pubmed: ROUGE SU4-Recall}
\includegraphics[width=0.85\textwidth,height=\textheight,keepaspectratio]{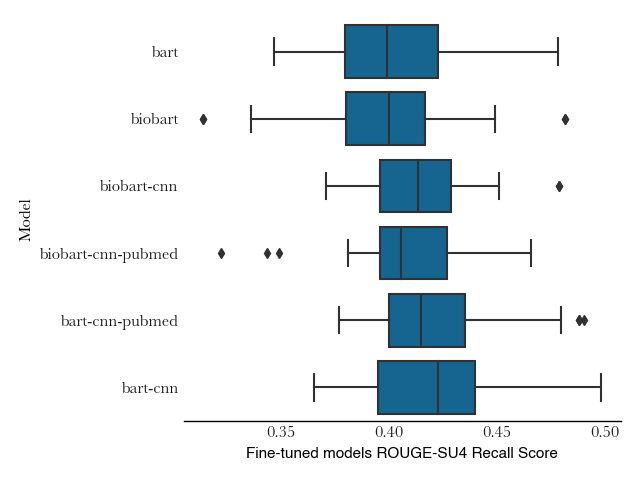}
\end{figure*}

\end{document}

%% file: sections/1-introduction-v1.tex
%
\section{Introduction}

The fields of question-answering and summarization have witnessed significant advancements in recent years, with a shift from classification-based extractive approaches to the emergence of abstractive summarization models. 
This transition has been driven by the superior performance and enhanced generalization capabilities exhibited by abstractive models, effectively blurring the boundary between long-form question answering and summarization. 
This paper addresses the summarization challenge presented by BioASQ Task B Phase B in the biomedical domain, for which we propose a novel approach.

The healthcare sector holds immense potential for leveraging health research data sharing to enhance clinical care, informed decision-making, and scientific discovery~\cite{Krumholz2016}. 
Sharing biomedical and healthcare studies and research data with the wider public requires robust and efficient methods. 
\emph{Large pre-trained language models} (LLMs) have emerged as promising candidates for this purpose. 
LLMs have the potential to store medical knowledge while accommodating variations in data and application tasks \cite{Bommasani2021}.
This paper aims to analyze the impact of the training process on LLMs' ability to store biomedical knowledge, explicitly focusing on their utilization for a question-answering and summarization task.

\begin{figure*}[t]
  \centering
  \includegraphics[width=0.7\textwidth,height=\textheight,keepaspectratio]{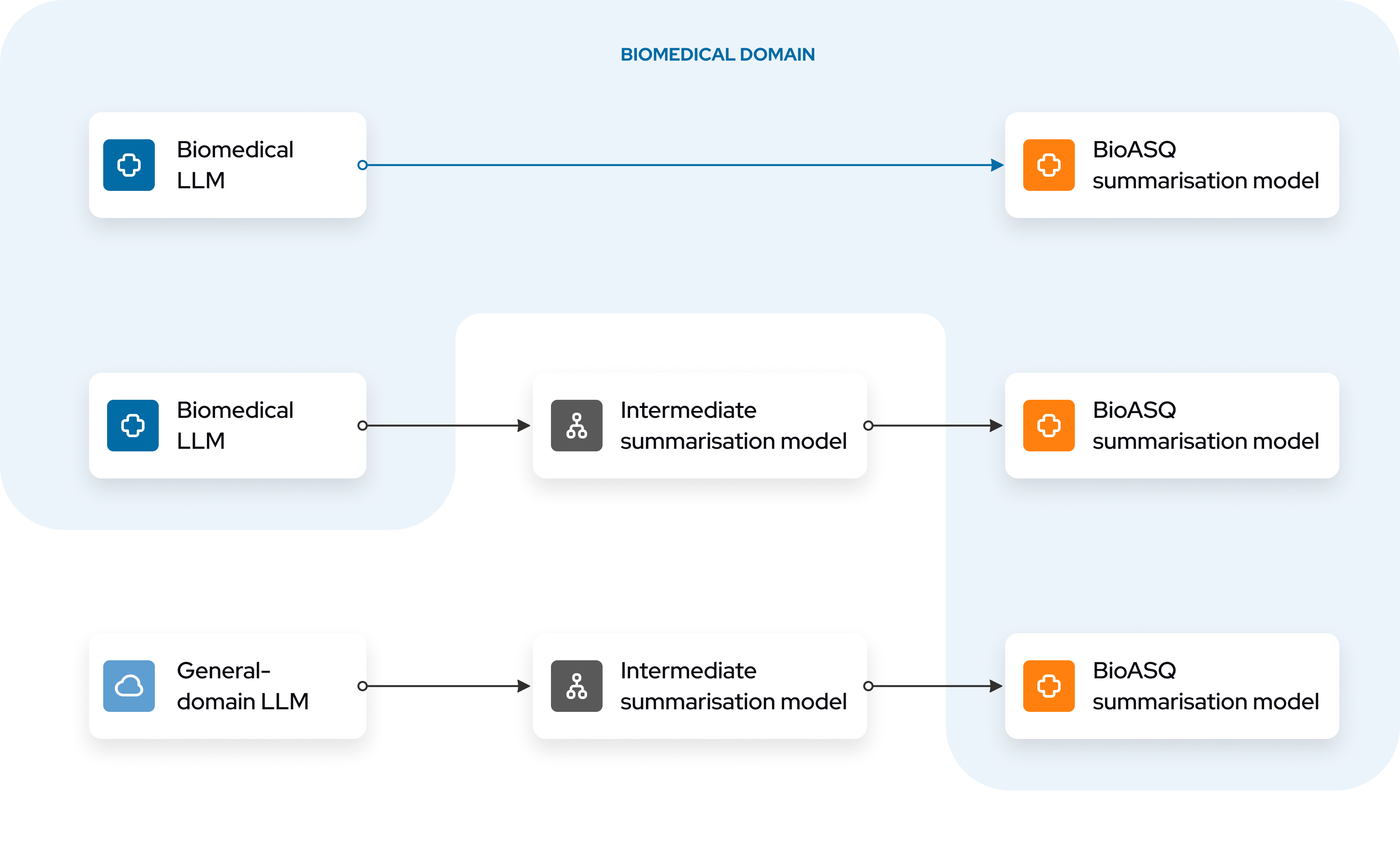}
  \caption{
    \textbf{Schematic representation of LLM usage in practice.}
    The blue arrow in the top row represents the two-phase (training and then fine-tuning) transfer learning approach typically seen in literature.
    The middle and bottom rows represent the task-specific fine-tuning approach proposed in this paper. 
    Our results suggest that in-domain pre-training does not improve the BioASQ summarization performance even when summarization training is introduced in the middle (second row). 
    In this diagram, the orange area indicates in-domain training.
  }
  \label{figure:tl-diagram}
\end{figure*}

Traditionally, achieving state-of-the-art performance on natural language processing tasks involves a two-phase approach \cite{Vaswani2017, Devlin2018} that is shown in blue in the top row of \cref{figure:tl-diagram}:
pre-training the models on an extensive range of texts and topics, followed by task-specific fine-tuning \cite{Devlin2018,Lewis2019,Bommasani2021}. 
This approach has revolutionized various areas of natural language processing \cite{mikolov13,Peters2018,Devlin2018}, with LLMs such as BERT, GPT, and BART demonstrating remarkable capabilities. 
However, pre-training models is a time-consuming and a resource-intensive process, and the literature lacks comprehensive insights into the performance of these models for domain-specific applications with limited data availability. 
Therefore, this study aims to address this gap by examining the performance of LLMs in the context of the BioASQ summarization task.

This paper investigates two open questions concerning biomedical domain question-answering and text summarization tasks.
Over the past five years, the biomedical domain has increasingly relied on in-domain pre-training and fine-tuning of BERT \cite{Devlin2018} for a wide range of datasets and benchmarks \cite{Alsentzer2019,Lee2019a,Gu2020,Alrowili2021,Raza2022}. 
In-domain pre-training has proven effective in enhancing performance for discriminatory biomedical tasks. 
However, BERT's architecture is not optimized for text generation tasks \cite{Liu2019b, Raffel2019}, lacking an autoregressive decoder to generate tokens based on previously generated ones. 
Consequently, BERT is suboptimal for generation tasks, necessitating exploring alternative approaches.
Previous studies evaluating biomedical models across diverse tasks have not reported results on generation problems due to using non-autoregressive models \cite{Gu2020}.
The first question is \textbf{is there a better-suited architecture for biomedical text generation tasks?}
A significant amount of research suggests that domain-specific pre-training significantly outperforms mixed-domain pre-training. 
However, we could not find any convincing evidence for supporting this belief when it comes to text generation problems \cite{Alsentzer2019,Lee2019a,Gu2020,Alrowili2021,Yuan2022,Luo2022,Raza2022}.
The second question is \textbf{do LLMs need to be pre-trained in domain to achieve optimal performance?}

We answer the above two questions.
To investigate the efficacy of domain-specific pre-training and fine-tuning for biomedical text generation, we propose an alternative three-step approach (shown in the bottom row of \cref{figure:tl-diagram}). 
In this approach, we initially train a general-domain LLM, followed by fine-tuning for a specific task in the general domain (text summarization) and subsequent fine-tuning for the target biomedical domain task. 
Contrary to established theories in the biomedical domain \cite{Gu2020, Yuan2022, Luo2022}, our findings suggest that having a large task-specific dataset can be more valuable than domain-specific pre-training for biomedical text generation tasks. 
This approach aligns with studies indicating that diverse pre-training objectives, larger and more diverse datasets, and tasks contribute to the robustness of the fine-tuning process even without domain adaptation \cite{Raffel2019, Hendrycks2020}.
 
 We explore alternative architectures for biomedical text generation.
 In this study, we focus on BART \cite{Lewis2019}, a comprehensive architecture that incorporates pre-training objectives from both BERT \cite{Devlin2018} and GPT \cite{Radford2018} models. 
 BART has demonstrated state-of-the-art performance in abstractive dialogue, question-answering, and summarization tasks, making it particularly effective for text generation and comprehension. 
 Our experimental results showcase the benefits and effectiveness of utilizing the BART architecture for transfer learning techniques in a context of a biomedical summarization task.
 
 \textbf{The main contributions of this work can be summarized as follows:}

\begin{itemize} 
\item Evaluating the advantages of domain-specific pre-training in the context of text generation tasks.
\item Evaluating the impact of task-specific training on improving text generation tasks.
\item Assessing the performance of BART, an encoder with an auto-regressive decoder architecture, in the biomedical question answering task B of BioASQ 11.
\end{itemize}